\pdfoutput=1
\documentclass{article}

\PassOptionsToPackage{square,sort,comma,numbers}{natbib}

\usepackage[final]{nips_2018}

\usepackage{authblk}
\usepackage[utf8]{inputenc} 
\usepackage[T1]{fontenc}    
\usepackage{hyperref}       
\usepackage{url}            
\usepackage{booktabs}       
\usepackage{amsfonts}       
\usepackage{nicefrac}       
\usepackage{microtype}      
\usepackage[tight,footnotesize]{subfigure}
\usepackage{graphicx}
\usepackage{subfigure}
\usepackage{caption}
\usepackage[usenames, dvipsnames]{color}
 
\graphicspath{{images/}}
\DeclareGraphicsExtensions{.pdf,.jpeg,.png}

\title{CheMixNet: Mixed DNN Architectures for Predicting Chemical Properties using Multiple Molecular Representations}

\author{Arindam Paul}
\author{Dipendra Jha}
\author{Reda Al-Bahrani}
\author{Wei-keng Liao}
\author{Alok Choudhary}
\author{Ankit Agrawal}
\affil{Department of Electrical Engineering and Computer Science, Northwestern University}

\begin{document}

\maketitle
\begin{abstract}

SMILES is a linear representation of chemical structures which encodes the connection table, and the stereochemistry of a molecule as a line of text with a grammar structure denoting atoms, bonds, rings and chains, and this information can be used to predict chemical properties. Molecular fingerprints are representations of chemical structures, successfully used in similarity search, clustering, classification, drug discovery, and virtual screening and are a standard and computationally efficient abstract representation where structural features are represented as a bit string. Both SMILES and molecular fingerprints are different representations for describing the structure of a molecule.
There exist several predictive models for learning chemical properties based on either SMILES or molecular fingerprints.
Here, our goal is to build predictive models that can leverage both these molecular representations.
In this work, we present CheMixNet- a set of neural networks for predicting chemical properties from a mixture of features learned from the two molecular representations - SMILES as sequences and molecular fingerprints as vector inputs. We demonstrate the efficacy of CheMixNet architectures by evaluating on six different datasets.
The proposed CheMixNet models not only outperforms the candidate neural architectures such as contemporary fully connected networks that uses molecular fingerprints and 1-D CNN and RNN models trained SMILES sequences, but also other state-of-the-art architectures such as Chemception and Molecular Graph Convolutions. 
\end{abstract}
\section{Introduction}
\label{sec:intro}
Traditionally, chemists and materials scientists have relied on experimentally generated or simulation-based computational data to discover new materials and understand their characteristics. The slow pace of development and deployment of new/improved materials has been considered as the main bottleneck in the innovation cycles of most emerging technologies~\cite{kalidindi2015data}. Data-driven techniques provide faster methods to identify important properties of chemical compounds and to predict feasibility to synthesize in chemical laboratories and thus promise to accelerate the research process of new materials development. There have been many initiatives to computationally assist molecular and materials discovery using machine learning (ML) techniques~\cite{kuz2011interpretation,agrawal2016perspective, yao2004comparative, gagorik2017improved, meredig2014combinatorial, ward2016general, agrawal2014exploration, mojitaba2018}. Conventional machine learning approaches for predicting chemical properties have emphasized the importance of leveraging domain knowledge when designing model inputs. Current research has demonstrated that deep neural networks (DNNs) have generally outperformed traditional machine learning models. DNN models are capable of learning representations, which sets it apart from conventional ML algorithms used in chemistry. Representation learning is the process of transforming input data into a set of features that can be effectively exploited to identify patterns from the data. In the context of chemistry, the analogous process would be to use deep learning (DL) to examine chemical structures and to construct features similar to engineered chemical features, with minimal assistance from an expert chemist. This approach that leverages representation learning of deep neural networks is a significant departure from the traditional research paradigm in chemistry.

In this work, we develop CheMixNet- a set of neural networks for predicting chemical properties by leveraging multiple molecular representations as inputs. We used simplified molecular-input line-entry system (SMILES)~\cite{toolkit1997daylight} notations as sequence inputs and molecular fingerprints as vector inputs. SMILES is a line notation of chemical structures which encodes the connection table and the stereochemistry of a molecule as a line of text. Our work improves upon the existing state-of-the-art approach of directly learning from vector representations such as molecular fingerprints or chemical text representations such as SMILES by harnessing the network structure of both forms of representations. CheMixNet is a variation of multi-input-single-output (MISO)~\cite{abdullah2011multiple} architectures that learn the chemical properties from a mix of intermediate features learned from two different input representations - a vector input in the form of molecular fingerprints and a sequence input in the form of SMILES strings. In our experiments, we used MACCS fingerprints - a first 2D representation of chemical structure using 167 features. Although MACCS usually perform worse than other molecular fingerprints, we chose MACCS because of its simplicity and ease of interpretation. We perform significant experimentation to determine the best neural network structure for the CheMixNet architectures. 

We evaluated the effectiveness of our mixed approach for building DNN architectures by training CheMixNet on six different datasets- a large dataset composed of 2.3 million samples from the Harvard Clean Energy Project (CEP) database and five other relatively smaller datasets from the MoleculeNet~\cite{wu2018moleculenet,GLambard98} benchmark. 
Compared to other DL models, CheMixNet architecture outperforms fully connected MLP models trained on molecular fingerprints, recurrent neural networks (RNN) and 1-dimensional convolutional neural network (CNN) models trained on SMILES, as well as other models - convolutional molecular graphs (ConvGraph)~\cite{duvenaud2015convolutional} and Chemception~\cite{goh2017chemception}. 
For instance, we achieved a mean absolute percentage error (MAPE) of 0.24 \% on the CEP dataset; this is significantly better than the MAPE of 0.43 \% using CNN-RNN model.
The CheMixNet architectures, as well as the benchmark models, are made accessible for the research community at https://github.com/paularindam/CheMixNet~\cite{chemixnet}. 

\section{Background and Related Works}
\label{sec:background}
In this section, we present a description of the two molecular representations we use in this work - SMILES and molecular fingerprints, 
and discuss existing deep neural architectures for predictive modeling of chemical properties in the Quantitative structure-activity relationship (QSAR)/Quantitative structure-property relationship (QSPR)~\cite{karelson1996quantum} modeling.

\subsection{SMILES \& Fingerprints}
Line notations are linear representations of chemical structures which encode the connection table and the stereochemistry of a molecule as a line of text ~\cite{warr2011representation}. 
SMILES~\cite{toolkit1997daylight} is the most popular specification in the form of a line notation to describe the structure of chemical species using short ASCII strings encoding molecular structures and specific instances.
One or more organic molecules attach to form long continuous chains known as branches.  SMILES has a grammar structure in which alphabets denote atoms,  special characters such as = and  $\equiv$ bond denote the type of bonds, encapsulated numbers indicate rings, and parentheses represent side chains. In this work, we limit ourselves to character level representation and do not explicitly encode the grammar. 

Molecular fingerprints are representations of chemical structures, successfully used in similarity search~\cite{johnson1990concepts}, clustering~\cite{mcgregor1997clustering}, classifications~\cite{breiman1984classification}, drug discovery~\cite{vass2016molecular}, and virtual screening~\cite{willett2006similarity}, a standard and computationally efficient abstract representation where structural features are represented by either bits in a bit string or counts in a count vector. Fingerprints were motivated by the need to find materials that match target material properties. They follow the assumptions that the properties of the material is a direct function of its structure and that materials with similar structure are likely to have similar physical-chemical character. Different fingerprints represent different aspects of a molecule, and thus each type of fingerprint can have different suitability for mapping to particular physical property. Various machine learning (ML) algorithms have been used to predict the activity or property of chemicals using molecular descriptors and/or fingerprints as input features. 
In our experiments, we used MACCS fingerprints~\cite{durant2002reoptimization,keys2005mdl} - a primitive 2D representation of chemical structure using 167 features. MACCS fingerprints were originally developed for the purpose of substructure screening. Unlike other hashed 1024 bit fingerprint representations such as Atom Pair and Topological Torsion that are difficult to comprehend, MACCS fingerprints represent the counts of the presence or absence of chemical fragments, and are easily comprehensible with each key having its own definition (e.g. key 99 indicates if there is a C=C bond, key 165 indicating if there is a ring present, key 125 representing if there are more than one aromatic rings in the structure).

\subsection{Related Works}

In their SMILES2vec~\cite{goh2017smiles2vec} paper, Goh et al. developed a RNN neural network architecture trained on SMILES for predicting chemical property. 
SMILES2vec was inspired by language translation using RNN. 
Goh et al. did not explicitly encode information about the grammar of SMILES. Instead, they anticipate  RNN units to learn these patterns implicitly and develop intermediate features that would be useful for predicting a variety of chemical properties. RNNs, particularly those based on LSTMs~\cite{hochreiter1997long} or GRUs~\cite{hochreiter1997long} are effective neural network designs for learning from text data. Their effectiveness has been demonstrated in multiple works such as the Google Neural Translation Machine that uses an architecture of 8+8 layers of residual LSTM unit~\cite{wu2016google}. In SMILES2vec, they modeled sequence-to-vector predictions, where the sequence is a SMILES string, and the vector is a measured chemical property. As SMILES is a chemical language and different from spoken language, commonly-used techniques in natural language processing (NLP) research, embeddings such as Word2vec~\cite{mikolov2013efficient} cannot be directly applied. 
In addition, they explored the utility of adding a 1D convolutional layer between the embedding and GRU/LSTM layers.
Goh et al.~\cite{goh2017chemception} developed “Chemception”, a deep CNN for the prediction of chemical properties, using only the images of 2D drawings of molecules. It was inspired by Google’s Inception-ResNet~\cite{szegedy2017inception} deep CNN for image classification. They utilized “raw data” in the form of 2D drawings of molecules that requires the minimal amount of chemical knowledge to create, and investigated the viability of augmenting and possibly eliminating human-expert feature engineering in specific computational chemistry applications.  Chemception was developed based on the Inception-ResNet v2 neural network architecture that combines arguably the two most important architecture advances in CNN design since the debut of AlexNet~\cite{krizhevsky2014imagenet} in 2012 - Inception modules and deep residual learning. During the training of Chemception,  additional real-time data augmentation to the image was performed so as to bolster the limited number of data available for each task. 
Finally, fully connected (MLP) architectures trained on fingerprint representations~\cite{ma2015deep, pyzer2015learning, ramsundar2015massively} are very popular in the cheminformatics community for predicting chemical properties. Although, MLP architectures trained on fingerprints are one of the earliest applications of deep learning in QSAR/QSPR modeling, they have consistently outperformed traditional ML models such as random forest and logistic regression, and DNN methods such as convolutional molecular graphs.

\section{Method}
\label{sec:method}
\subsection{Motivation}
Several works~\cite{kuz2011interpretation, kan2015improving} have demonstrated the effectiveness of ensemble of different kinds of neural networks for improvement in model performance over the individual candidate neural networks. Fully connected deep neural network architectures trained on fingerprint representations~\cite{ma2015deep, pyzer2015learning, ramsundar2015massively} are very popular in the cheminformatics community for predicting chemical properties. Goh et al.~\cite{goh2018smiles2vec} propounded the SMILES2vec architecture for treating SMILES strings as text sequences and trained recurrent neural network architectures. SMILES2vec and MLP architectures have been among the most successful neural network architectures in predicting chemical properties. In this paper, we harness the efficacy of these architectures and mix them into one architecture which we refer as CheMixNet. SMILES and fingerprints are the two most common representations of chemical molecules. By allowing a neural network to learn from both these representations, we could increase the generalizability and the scope of the architecture. 
Sequence classification on shorter texts is generally harder than on longer texts and usually has worse performance than longer texts. As SMILES2vec essentially treats the SMILES as a text with character level embedding, the performance of SMILES2vec degrades on shorter strings. Also, the performance of MLP models trained on molecular fingerprints generally varies based on the size of the molecule (performance varies based on small versus large organic molecules). CheMixNet provides a model architecture that can leverage the best of both forms of representation learned from the two inputs using appropriate neural network components for them. This provides the network with the ability to automatically assess the degree to which each representation can be leveraged for learning the given chemical property.

\begin{figure}[]
\centering
\includegraphics[width = 5.5in, keepaspectratio=true]{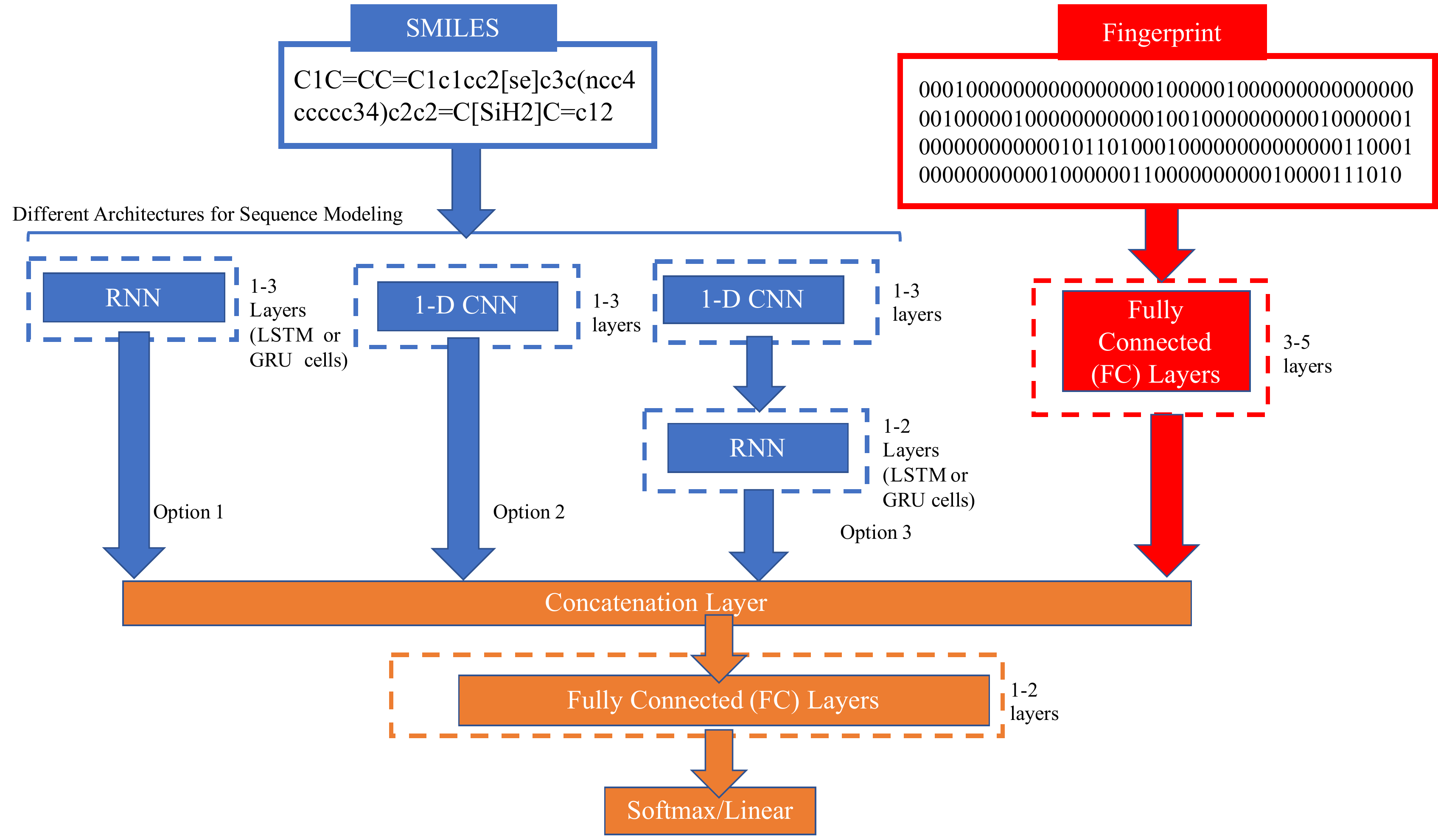}
\caption{The proposed CheMixNet architecture for learning from the two molecular representations. The blue branch represents candidate neural networks for learning from SMILES sequences. Option 1 uses only LSTMs/GRUs for modeling the SMILES sequences, option 2 uses only 1-D CNNs for sequence modeling, and option 3 uses 1-D CNNs followed by LSTMs/GRUs. The fully connected (FC) branch of the model with molecular fingerprint inputs is illustrated in red. The orange part represents the fully connected layers that learn the final output from the mix of intermediate features learned by the two network branches. We exemplify the molecular fingerprint and SMILES with one representative example in this illustration. }
\label{fig:model}
\end{figure} 

\subsection{Design}
Figure~\ref{fig:model} illustrates our design approach for building CheMixNet models.
We present three different architectures where we mix the output features learned using different types of models to learn the chemical properties from the two molecular representations as inputs- the fingerprints and the SMILES formula; hence, referred as CheMixNet.
They are basically composed of two neural network branches - a sequence modeling branch that learns from the SMILES sequences using 1-D CNN and/or RNN, and a fully connected (FC) branch that learns from the MACCS fingerprint representation.
The two input representations are shown in the top of Figure~\ref{fig:model}.
Since the SMILES representations are composed of a sequence of characters, the first network branch for learning from SMILES are composed of one of the following: 1) RNN, 2) 1-D CNN, and 3) 1-D CNN followed by RNN.
Since the fingerprints are composed of bit or count vectors, the second branch for learning from them are composed of multiple layers of fully connected layers.
The individual models that leverage a single input representation are composed of one of these components along with some fully connected layers at the end.

To build the mixed DNN architectures, we combine the intermediate features learned using the two branches and applied some fully-connected layers for learning the final regression or classification output.
Depending on the learning task, the output layer uses sigmoid activation for the binary classification tasks or linear activation for the regression tasks. 
Since we have three candidate neural networks for sequence modeling using SMILES, ChemixNet contains three neural network architectures which we refer as CNN-FC, RNN*FC, and CNN-RNN*FC where the `-' represents networks stacked in sequence and `*' represents the combination of intermediate features learned using two parallel network branches.

\section{Data}
\label{sec:data}
\subsection{Description of the Datasets}
We demonstrate the effectiveness of our approach for designing DNN architectures for learning chemical properties from SMILES and fingerprints using six different datasets as shown in Table~\ref{tab:Data}.
First, Harvard CEP Dataset~\cite{pyzer2015learning, CEPData} contains molecular structures and properties for 2.3 million candidate donor structures for organic photovoltaic cells. 
Organic Photovoltaic cells (OPVs)~\cite{sang2013dft,hu2012theoretical,mohamad2015first,inostroza2016improvement} are lightweight, flexible, inexpensive and more customizable compared to traditional silicon-based solar cells~\cite{hoth2009topographical}.  
For a solar cell, the most important property is power conversion efficiency or the amount of electricity which can be generated due to the interaction of electron donors and acceptors, which are dependent on the HOMO values of the donor molecules. 
In this work, we considered highest occupied molecular orbitals (HOMO) as the target property as it determines the power conversion efficiency of a solar cell according to the Scharber model~\cite{scharber2006design}.
Next, we used the Tox21, HIV, ESOL, and FreeSolv (Experimental and Computed) datasets from the MoleculeNet~\cite{wu2018moleculenet,GLambard98} benchmark repository; they involve two classification and two regression tasks. 
The Tox21 dataset is an NIH- funded public database of toxicity measurements comprising of 8981 compounds on 12 different measurements ranging from stress response pathways to nuclear receptors. This dataset provides a binary classification problem of labeling molecules as either “toxic” or “non-toxic”. 
The FreeSolv dataset is comprised of 643 compounds that have computed and experimental hydration free energies of small-molecules ranging from –25.5 to 3.4 kcal/mol; we refer to the dataset containing experimental values as FreeSolv-Exp and the one with computed property values as FreeSolv-Comp. Hydration free energy is a physical property of the molecule which can be computed from first principles. 
ESOL is a dataset containing 1128 compounds with water solubility (log solubility in mols per litre) for common organic small molecules. Lastly, we evaluated the performance of CheMixNet on the HIV dataset obtained from the Drug Therapeutics Program AIDS Antiviral Screen, which measured the ability of 41,913 compounds to inhibit HIV replication in vitro. Using the curation methodology adopted by MoleculeNet, this dataset was reduced to a binary classification problem of “active” and “inactive” compounds. The original HIV dataset was very imbalanced with the minority class comprising less than 4 \% of the dataset. In our work, we decided to balance the minority classes by under-sampling the majority class by randomly selecting 1,443 (size of samples from the minority class) samples. 
Although this led to a significant reduction in the size of the dataset, it also allowed us to investigate the viability of CheMixNet architecture on smaller datasets without significant network topology changes. 

 \begin{table}[]
\centering

\small{
\caption{Description of all the 5 datasets used to evaluate the performance of CheMixNet architectures. The original HIV dataset had 41,193 compounds but reduced to 2,886 after under-sampling.}
\label{tab:Data}
\begin{tabular}{|l|c|c|c|}
\hline
\textbf{Dataset} & \textbf{Property} & \textbf{Task}  & \textbf{Size}\\ \hline
CEP&Highest Occupied Molecular Orbital Energy&Regression&2,322,849 \\ \hline
HIV*&Activity&Classification&2,886 \\ \hline
Tox21&Toxicity&Classification&8,981 \\ \hline
FreeSolv-Exp (Experimental)&Solvation Energy &Regression&643 \\ \hline
FreeSolv-Comp (Computed)&Solvation Energy &Regression&643 \\ \hline
ESOL&Solubility&Regression&1,128 \\ \hline
\end{tabular}}
\end{table}

\subsection{Dataset Preparation}
For the SMILES sequence, we used 1-hot encoding to convert the SMILES into a fixed length representation. The length of the sequence was determined by the length of the longest SMILES sequence in each dataset. We applied zero padding for shorter strings so that we had a uniform sequence of size equal to the maximum length for each dataset. The vocabulary size was determined by finding the number of unique characters in each dataset. 
Table~\ref{tab:input} describes the size of vocabulary and the maximum input length for all datasets.
The datasets were randomly split in the ratio of 4:1 into training and test sets. Further, the training set was split into 9:1 ratio for training and validation.

 \begin{table}[]
\centering
\caption{Vocab size and Maximum Input length for the datasets }
\label{tab:input}
\begin{tabular}{|l|c|c|}
\hline
\textbf{Dataset} & \textbf{Size of Vocabulary} & \textbf{Maximum Input Sequence Length}\\ \hline
CEP&23&83 \\ \hline
HIV&54&400 \\ \hline
Tox21&42&940 \\ \hline
FreeSolv-Exp&32&83 \\ \hline
FreeSolv-Comp&32&83 \\ \hline
ESOL&33&98 \\ \hline
\end{tabular}
\end{table}

\section{Experiments \& Results}
\label{sec:results}
In this section, we present the experimental settings and results of the CheMixNet architectures and the comparison with other contemporary DL models on the 2.3 million CEP dataset as well as the five datasets from the MoleculeNet benchmark. 
\subsection{Experimental Settings} 
The DNN models were implemented using Python and Keras~\cite{chollet2015keras} with TensorFlow~\cite{abadi2016tensorflow} as the backend. They are trained using Adam as the optimization algorithm with a mini-batch size of 32. For generating the MACCS fingerprints, we used RDKIT~\cite{landrum2006rdkit} library. Scikit-Learn~\cite{pedregosa2011scikit} was used for data preprocessing and for evaluating the test set errors. All experiments are carried out using NVIDIA DIGITS DevBox with a Core i7-5930K 6 Core 3.5GHz desktop processor, 64GB DDR4 RAM, and 4 TITAN X GPUs with 12GB of memory per GPU. For our experiments, we used a learning rate of 0.001. 
We used the mean squared error (MSE) as the loss function for the regression tasks and used the mean absolute \%  error (MAPE) as the performance metric.
For classification tasks, we used the binary cross-entropy as the loss function and used the area under the ROC curve (AUC) as a performance metric.
Early stopping was used during training to avoid over-fitting. 
For the benchmark results for graph convolution networks, we used the DeepChem~\cite{deepchem} library. For benchmarking with Chemception, there is no official public library, so we implemented the network using Keras which is also available in the CheMixNet repository~\cite{chemixnet}.

We used the libraries hyperas~\cite{hyperas} and hyperopt~\cite{hyperopt} to perform Bayesian hyperparameter search~\cite{snoek2012practical} to obtain the best choice of layer depth (for MLPs, 1-D convolutional and LSTM/GRU units), number of recurrent units for LSTMs/GRUs and learning rate. 
Further, the Bayesian hyperparameter search was performed only for the CEP database.
Once we determined the best hyperparameters for CEP, we did not change any hyperparameter except the batch size.
For the CEP dataset, we used a batch size of 64; for the two classification datasets (HIV and Tox21), we used a batch size of 32; for the ESOL and FreeSolv, we used a batch size of 16. 

\subsection{Results}

We evaluated the effectiveness of our mixing approach for building DNN architectures to learn from both molecular representations by training the CheMixNet using six different datasets. To compare their performance, we also trained other state-of-the-art architectures for all datasets used.  This includes the fully connected (FC) networks trained on the MACCS fingerprints, 
the two broad classes of SMILES2vec architectures - RNN and CNN-RNN, novel experimentation on SMILES using 1-D convolutions, ConvGraph and Chemception model.
For the RNN and CNN-RNN architectures, we experimented using both LSTM and GRU units.
As previous works~\cite{zhang2015character, joulin2016bag} have demonstrated the efficacy of 1-D CNN to perform effectively in text prediction without any recurrent component, we compared against 1-D CNN trained on SMILES sequences. Lastly, we compare against ConvGraph architecture that uses the molecular structure encoded as graphs as input, and Chemception architecture that uses chemical images as input.

\subsubsection{Performance on the CEP Dataset}
\begin{figure}[]
\centering
\label{fig:barchart_results}
\subfigure[Comparison of training curves]{
\label{fig:hiv_auc}
\includegraphics[width=.43\textwidth]{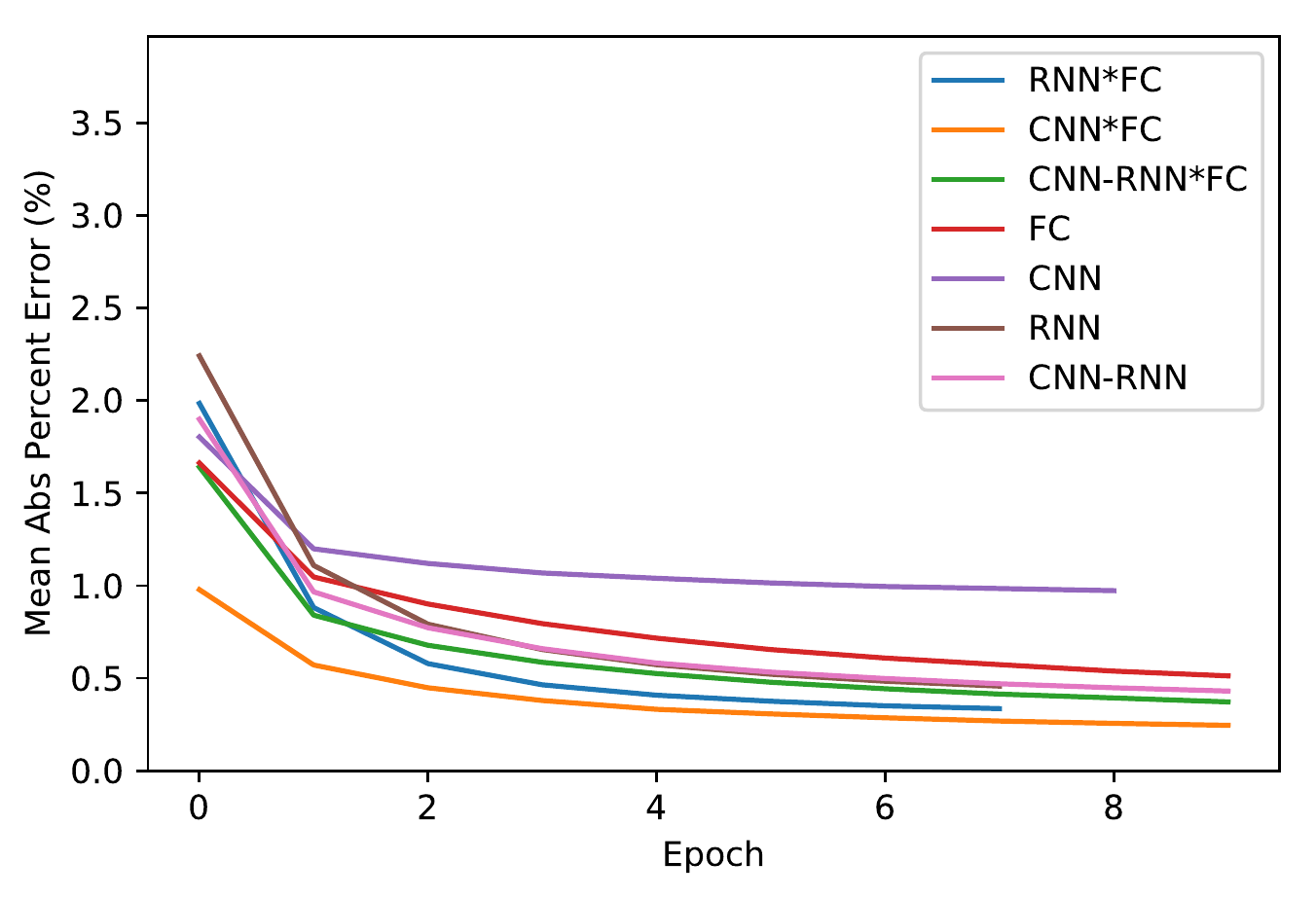}}
\qquad
\subfigure[MAPE comparison]{
\label{fig:tox_auc}
\includegraphics[width=.47\textwidth]{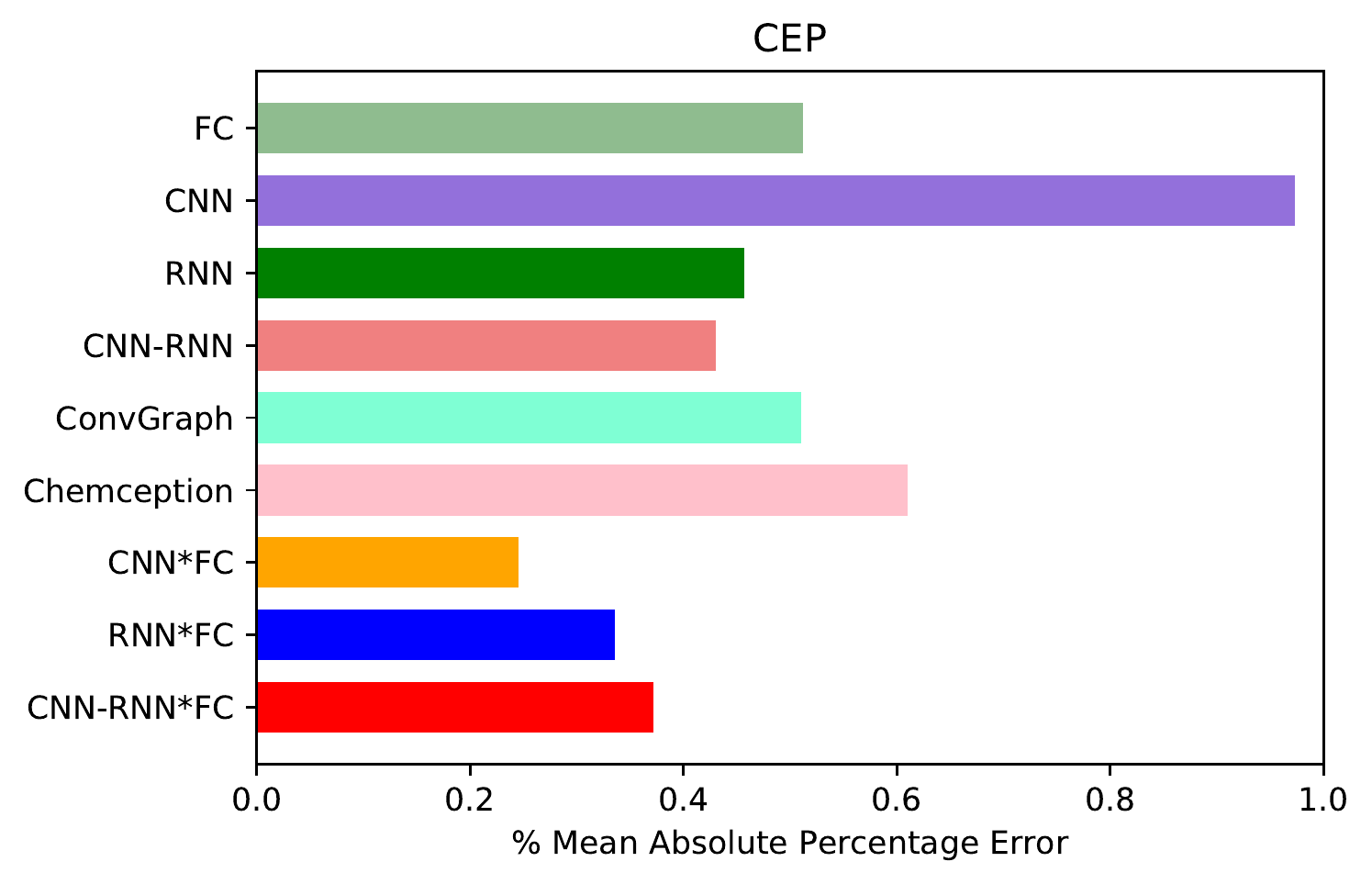}}
\caption{Comparison of the training error curves and mean absolute percentage error on the test set for different DNN architectures on the CEP dataset. The '-' sign indicates when the networks are trained in sequence and '*' when two parallel multi-input networks (one with SMILES as input and the other with fingerprints as input) are concatenated. 
In these experiments, we use MACCS fingerprints - however, metrics from other fingerprints were similar.
Our results demonstrate that the CNN*FC model performs the best. The three mixed networks perform comparatively better than the other state-of-the-art models.
Since we use ConvGraph module from deepchem repository out of the box which does not give any information about convergence while Chemception usually takes about 100 epochs to converge, the training curve for Chemception and ConvGraph is not shown. }
\label{fig:res_cep}
\end{figure}

Figure~\ref{fig:res_cep} demonstrates the performance results of different DNN models on the CEP dataset.
For the presented results, we used the MACCS fingerprints; similar metrics were observed using other types of fingerprints.
The existing models trained on the SMILES generally perform better than the models trained using only fingerprints;
CNN-RNN and RNN perform significantly better than the FC model.
We conjecture the difference in performance results from the difference in feature representation using SMILES and fingerprints.
Since fingerprints are generated from SMILES, fingerprints are supposed to contain less information.
We experimented using both LSTM and GRU for building the models composed of RNNs; for the CNN-RNN*FC, LSTM performs better than GRU while GRU performs best for CNN-RNN.
Our results illustrate that the three mixed networks perform comparatively better than the existing candidate model architectures; the CNN*FC model performing significantly better.
The CNN branch of CNN*FC model is composed of an embedding layer of length 32 followed by two 1-D convolutional layers with 32 filters with a kernel size of 3 (same padding and ReLU as the activation function).
The FC branch is composed of four fully connected layers with 1024, 512, 256 and 64 units respectively.
The final network that learns on the mixed intermediate features is composed of two layers with 64 and 1 outputs respectively.
Since we perform an architecture search for each network independent of other networks, the architecture configuration for the mixed networks are different from the individual networks that leverage one input representation; hence, this is different from the current model ensemble approach where the outputs from different trained networks are aggregated to predict the output.

The derivation of fingerprints from SMILES involves simple logic and computation.
However, we find that the mixing of intermediate features learned using the two network branches from the two molecular representations trained resulted in a significant gain in performance.
It demonstrates the effectiveness of CheMixNet architectures in learning from multiple types of feature representations for better performance.

\subsubsection{Performance on MoleculeNet Datasets}

We further analyzed the effectiveness of using mixed networks in learning from multiple inputs by evaluating on five datasets from the MoleculeNet benchmark.
Two of these datasets involves classification tasks while the rest involves regression tasks.
Figure 3 illustrates the performance results of different types of model architectures on these datasets.
For the two classification tasks, we observe that CNN*FC performs better than all other models. The other two mixed models CNN-RNN*FC and RNN*FC perform better than the existing models except for the FC model. FC model performs better than all other existing models on the classification tasks from the HIV and Tox21 datasets.
For the regression problems, we observe similar patterns- one of the mixed networks performing the best among all the networks.
For the two FreeSolv datasets, CNN-RNN*FC performs the best; there is no one single best model among the existing network that works best for both these datasets.
For the ESOL dataset, RNN*FC performs the best among all models; CNN model performing slightly worse.
In general, we always observe benefit in performance from using mixed networks which can learn from both inputs- SMILES, and fingerprints.
Since fingerprints are derived from SMILES, we conjecture the gain in performance not only comes from multiple inputs but also and more importantly from the use of different types of networks for different input representations.

\begin{figure}[h]
\centering
\label{fig:barchart_results}
\subfigure[HIV ROC AUC comparison]{
\label{fig:hiv_auc}
\includegraphics[width=2.5in]{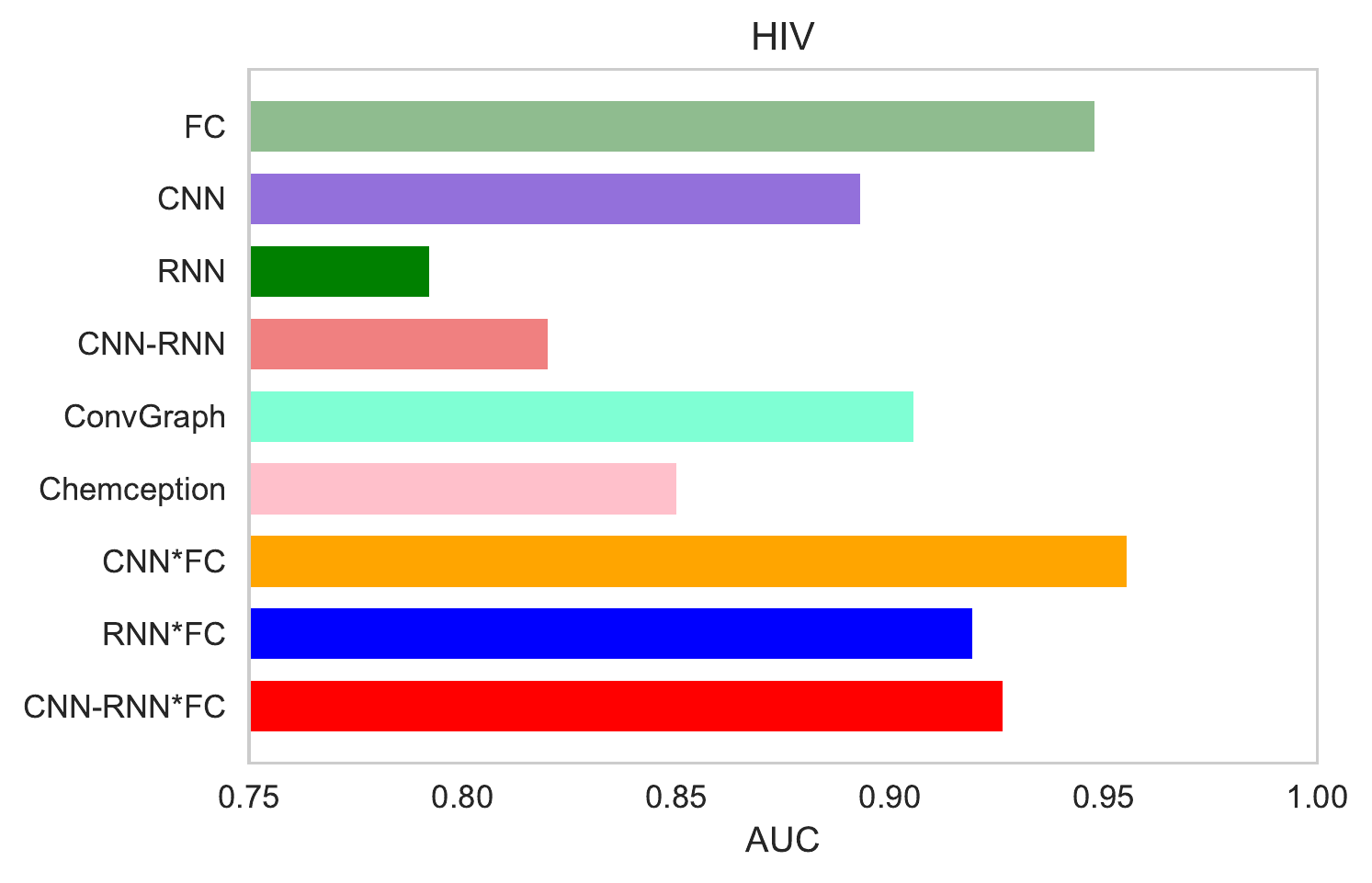}}
\qquad
\subfigure[Tox21 ROC AUC comparison]{
\label{fig:tox_auc}
\includegraphics[width=2.5in]{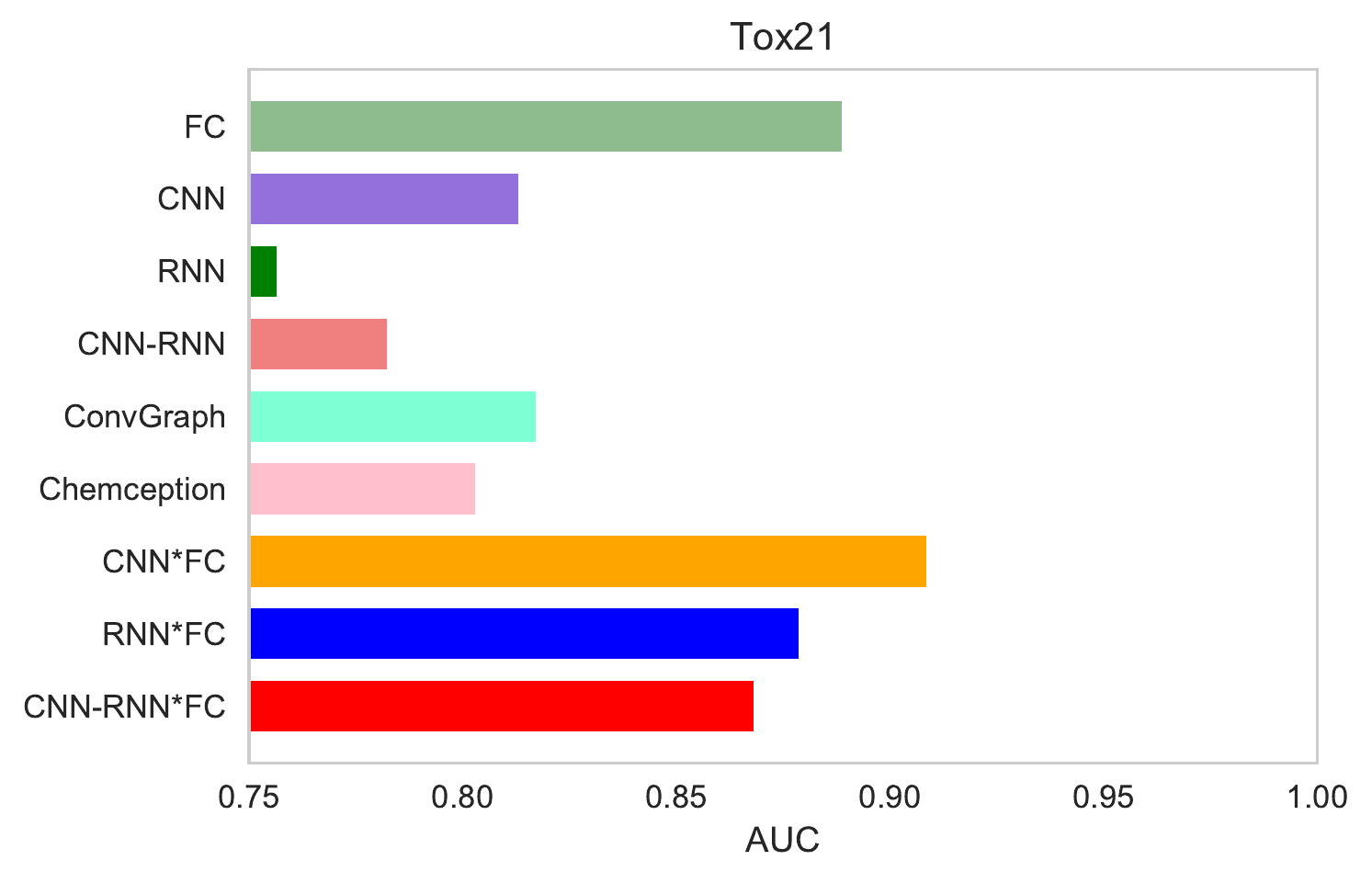}}
\qquad
\subfigure[FreeSolv-Comp MAPE comparison]{
\label{fig:freesolv_calc_mape}
\includegraphics[width=2.5in]{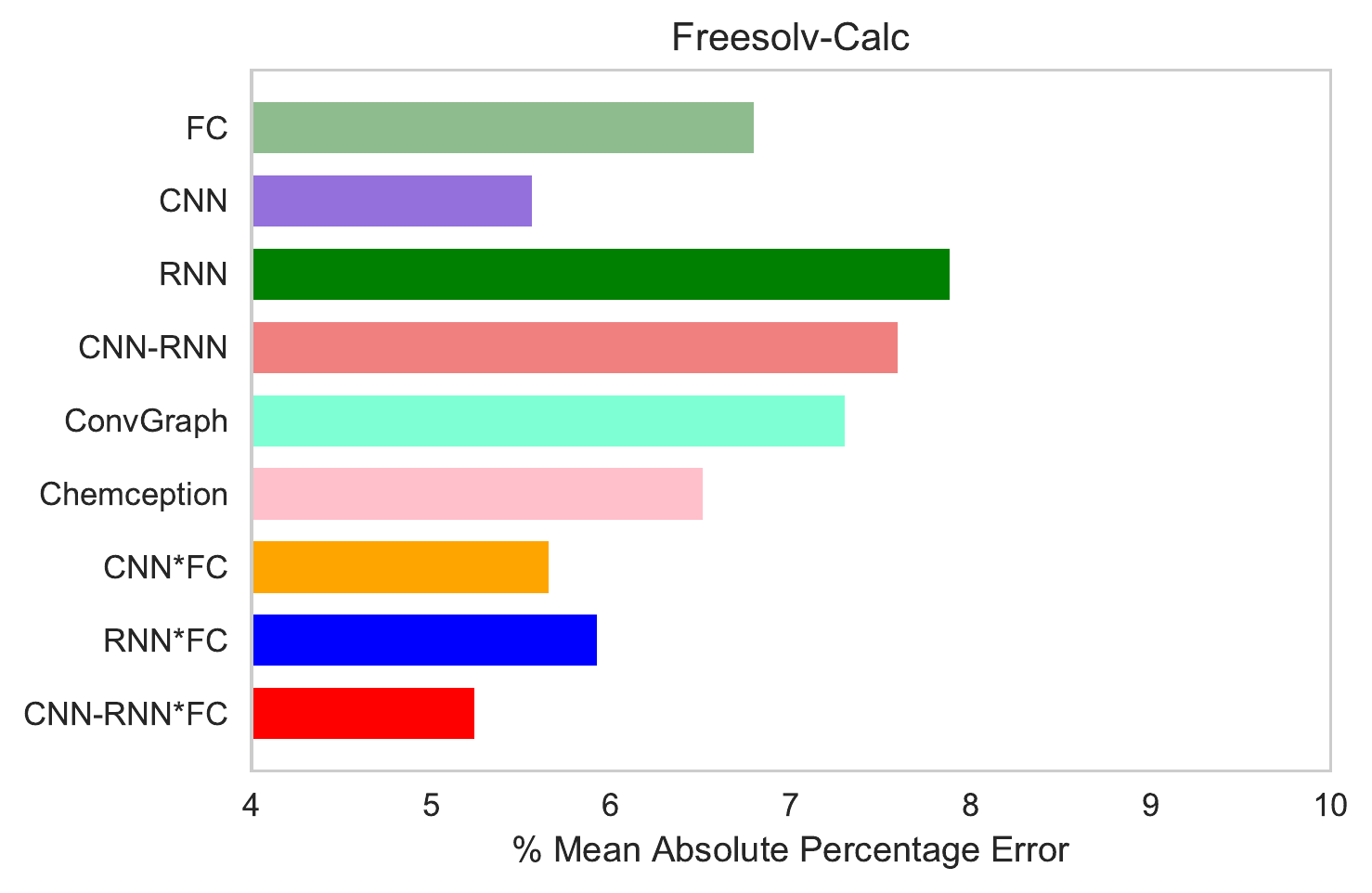}}
\qquad
\subfigure[FreeSolv-Exp MAPE comparison]{
\label{fig:freesolv_calc_mape}
\includegraphics[width=2.5in]{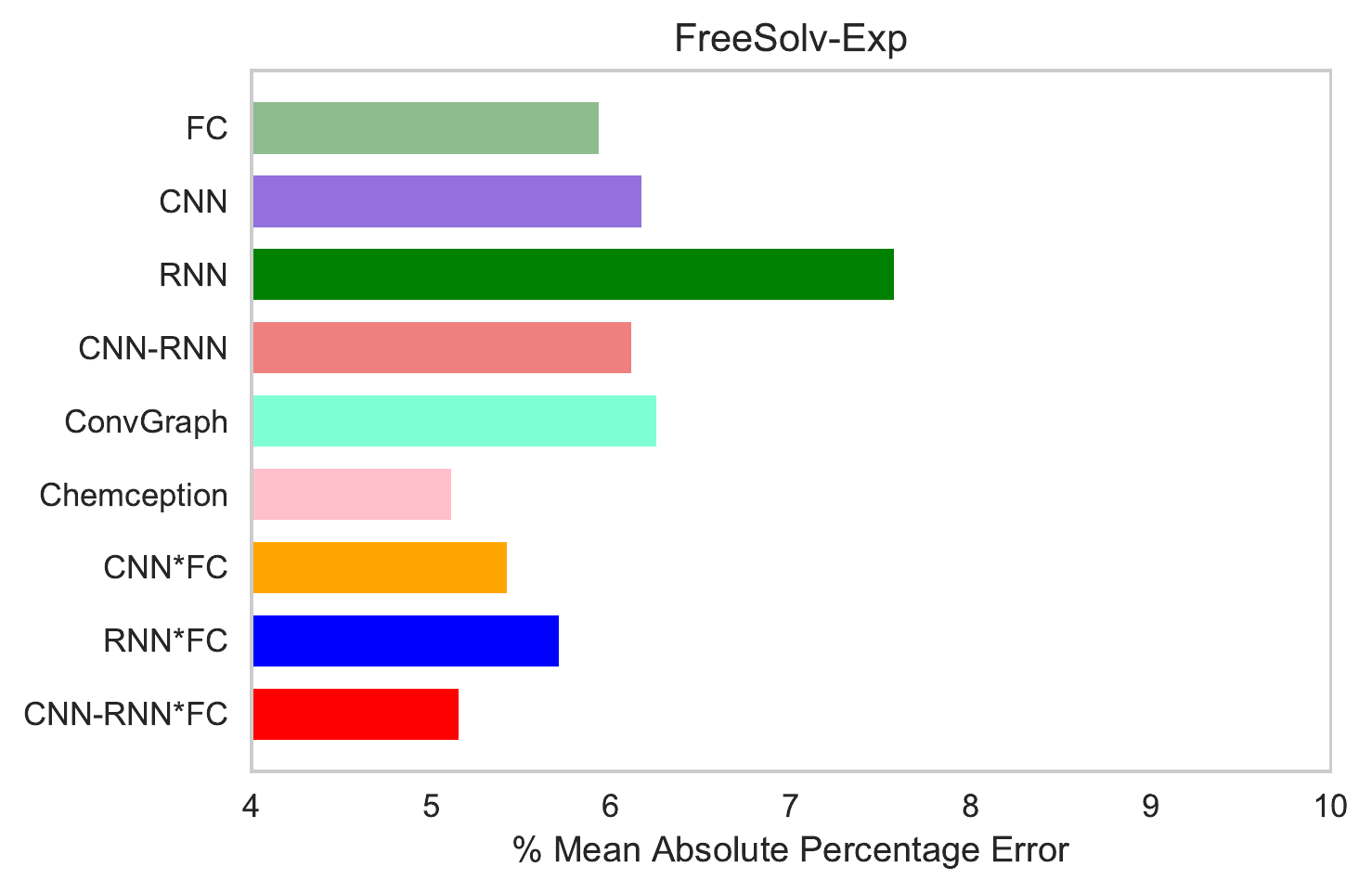}}
\qquad
\subfigure[ESOL MAPE comparison]{
\label{fig:esol_mape}
\includegraphics[width=2.5in]{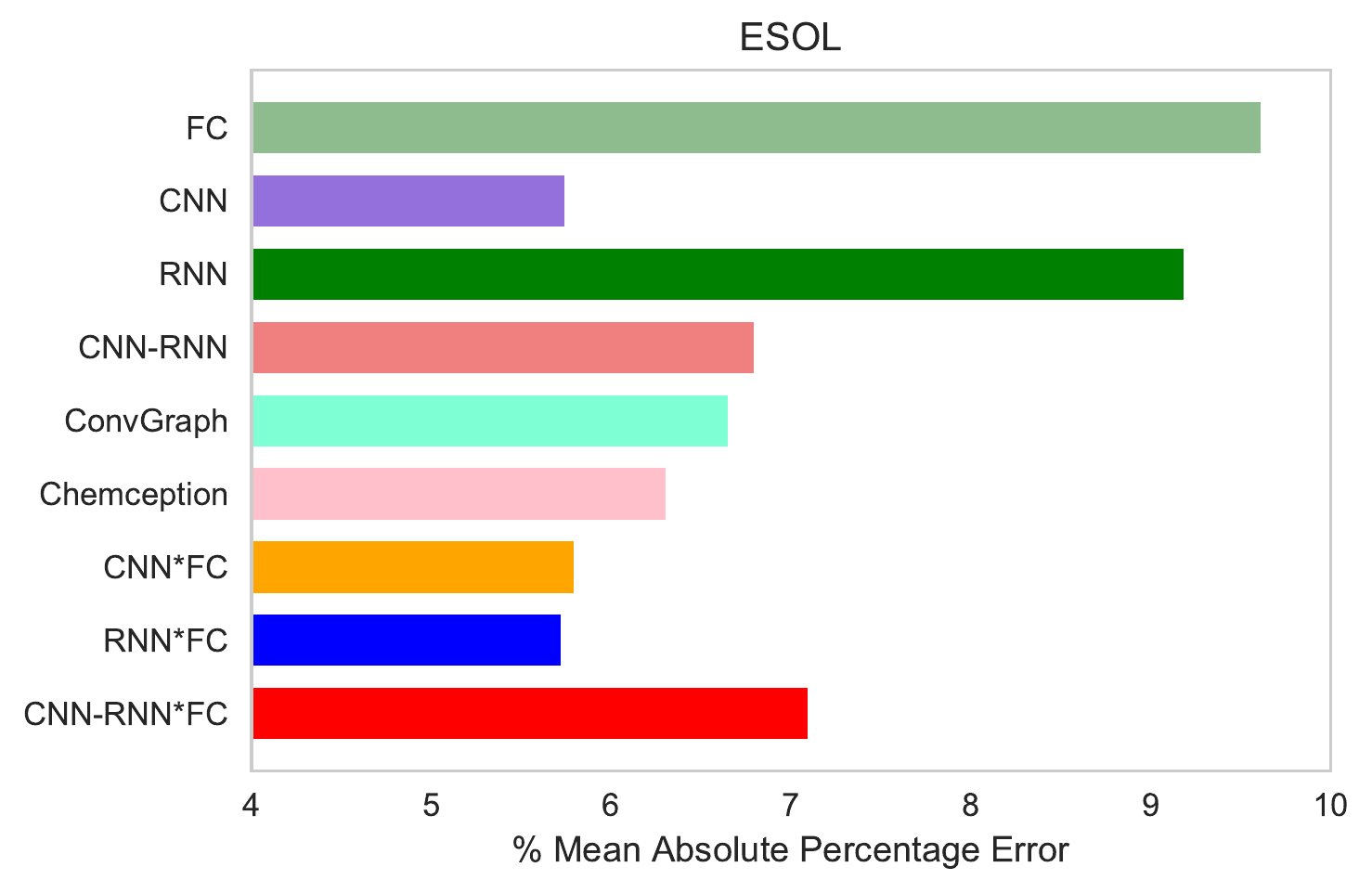}}
\label{res_other}
\caption{Performance of CheMixNet models against contemporary DNN models for the datasets from MoleculeNet benchmark. Tox21 and HIV datasets involve classification tasks while FreeSolv-Exp, FreeSolv-Comp and ESOL involve regression tasks; higher AUC is better for classification tasks while lower MAPE is better for regression tasks. CheMixNet outperforms the existing state-of-the-art models on all datasets. In general, the mixed models are better than the existing models.}
\end{figure}

\section{Conclusions and Future Work}
\label{sec:conclusions}
In this paper, we developed CheMixNet, the first mixed deep neural network that leverages both chemical text (SMILES) as well as molecular descriptors (MACCS fingerprints) for predicting chemical properties. 
Compared to existing DL models trained on single molecular representations, the proposed CheMixNet architectures perform significantly better on all the six datasets used in our study.
The results provide proof of concept of the efficacy of using mixed input architectures for chemical property prediction.  We demonstrate that by using a mixed deep learning approach, we can leverage the features of both sequence and fingerprint representations and achieve much better results, even with only a few hundred training samples. Further, the results demonstrate that CheMixNet architectures can be generalized over a different range of chemical properties independent of the type of supervised learning tasks (classification or regression) and the type and size of datasets. The range of chemical properties predicted in our study is relevant across solar cell technology, pharmaceuticals, biotechnology, and consumer products. The CheMixNet architectures, as well as the benchmark models, are made accessible for the research community at https://github.com/paularindam/CheMixNet~\cite{chemixnet}.

Although we limit the scope of our work in developing the CheMixNet architectures, 
we plan to extend this current work by trying to develop a method for interpreting the proposed mixed-input architecture as part of future work. 
For future CheMixNet architectures, we plan to explore ConvGraph and Chemception architectures as candidate input neural networks for the mixed neural network. 
Further, as chemical significance is present in both the character following as well as preceding a given character in a SMILES string, we believe bidirectional RNNs can perform better than vanilla one-directional RNNs. Lastly, we believe that Hierarchical Attention Networks (HANs)~\cite{yang2016hierarchical} that combine character level and word level sequences for text prediction could present superior performance to the aforementioned architectures.


\subsubsection*{Acknowledgments}

This work was performed under the following financial assistance award 70NANB14H012 from U.S. Department of Commerce, National Institute of Standards and Technology as part of the Center for Hierarchical Materials Design (CHiMaD). Partial support is also acknowledged from the following grants: NSF award CCF-1409601; DOE awards DE-SC0007456, DE-SC0014330; AFOSR award FA9550-12-1-0458.

\vfill
\bibliographystyle{unsrt}  
\bibliography{opv} 

 

\end{document}